\documentclass[runningheads]{llncs}

\usepackage{eccv}

\usepackage{eccvabbrv}

\usepackage{graphicx}
\usepackage{booktabs}
\usepackage{multirow} 

\usepackage[accsupp]{axessibility}  

\usepackage{hyperref}

\usepackage{orcidlink}

\begin{document}

\title{Adaptive Tversky Loss for Improved and Generalizable Medical Image Segmentation} 

\titlerunning{Adaptive Tversky Loss}


\authorrunning{Y. Yokoi and K. Hotta}

\author{
Yuto Yokoi\orcidlink{0009-0005-7404-8749}\inst{1}\and
Kazuhiro Hotta\orcidlink{0000-0002-5675-8713}\inst{1}
}

\authorrunning{Y. Yokoi and K. Hotta}

\institute{
Meijo University,Japan\\
\email{253427028@ccalumni.meijo-u.ac.jp}, \email{kazuhotta@meijo-u.ac.jp}
}

\maketitle

\begin{abstract}
Medical image segmentation remains challenging due to class imbalance and ambiguous boundaries, which hinder region extraction. Dice-based losses address these issues by measuring the overlap between predictions and ground truth. However, Tversky Loss requires manually tuned weights for false positives and false negatives, making its performance dependent on datasets and model architectures. This dependency can reduce generalizability and reproducibility. We propose an adaptive Tversky Loss that dynamically determines these weights from the Dice coefficient, eliminating manual hyperparameter tuning and reducing dataset- and model-dependent bias. We further introduce a multi-scale boundary-adaptive mechanism that incorporates boundary information to encourage progressive learning from coarse regions to fine boundary details. Experiments on medical image datasets and segmentation models show that our method outperforms conventional Dice-based losses and Tversky Loss overall, while maintaining stable performance across datasets and architectures. These results demonstrate its potential as a bias-reduced, generalizable, and trustworthy loss function for medical image segmentation.

\end{abstract}

\section{Introduction}
\label{sec:intro}
Medical image segmentation is a technique for extracting organs and lesion regions at the pixel level from medical images such as CT, MRI, ultrasound, and endoscopic images. It plays an important role in a wide range of clinical applications, including computer-aided diagnosis, treatment planning, and follow-up observation\cite{medicalsindan}. In recent years, the performance of medical image segmentation has been significantly improved by advances in deep learning models, including U-Net\cite{Unet} and its variants\cite{transunet,attentionunet,unetplusplus,unet3plus}. However, in medical images, target regions are often much smaller than background regions, resulting in severe class imbalance. In addition, the boundaries between lesions and normal tissues are often ambiguous, and the shape and intensity distribution of target regions can vary substantially due to imaging conditions, noise, and individual differences. Therefore, achieving accurate and stable segmentation remains an important challenge.

To address these challenges, the design of loss functions is a crucial factor that directly affects segmentation performance\cite{segloss}. Cross Entropy Loss, which minimizes pixel-wise classification errors, has been widely used. However, in medical images where background pixels dominate, the learning process may be biased toward the background region, leading to degraded extraction performance for small lesion regions. Therefore, Dice Loss\cite{dicelosssankou}, which directly evaluates the overlap between predicted and ground-truth regions, has been widely adopted. Although Dice Loss is relatively robust to class imbalance, it treats false positives and false negatives equally, making it difficult to flexibly reflect task-specific requirements such as reducing missed detections or suppressing over-segmentation.

On the other hand, Tversky Loss\cite{tversky} is a generalized form of Dice Loss that assigns different weights to false positives and false negatives, thereby controlling the balance between sensitivity and over-segmentation. Thus, depending on the characteristics of the task, it can emphasize the suppression of either false negatives or false positives and has the potential to achieve higher segmentation performance than Dice Loss\cite{focaltversk}. However, its performance strongly depends on the weighting coefficients, and the optimal settings vary according to the characteristics of the dataset and the model architecture. Therefore, to effectively use Tversky Loss, these weighting coefficients must be carefully tuned for each dataset and model. This tuning process imposes a practical burden and can be a factor that hinders the general use of Tversky Loss compared with the more easily applicable Dice Loss\cite{segloss}. Furthermore, weighting settings that are overly adapted to a specific data distribution or model architecture may reduce generalizability and reproducibility across different datasets and models.

To address these limitations, we propose Adaptive Tversky Loss (ATL), which dynamically determines the weights for false positives and false negatives based on the Dice coefficient. By updating the weights according to the prediction performance during training, ATL eliminates the need for manual tuning of the weighting coefficients required in conventional Tversky Loss, thereby mitigating dataset- and model-dependent weight settings and improving generalizability and reproducibility. We further propose Multi-Scale Boundary-Adaptive Tversky Loss (MS-BATL), an extension of ATL that incorporates boundary information\cite{boundary} through a multi-scale boundary-adaptive mechanism based on morphological operations with multiple kernel sizes. MS-BATL is designed to incorporate information ranging from coarse region-level structures to fine boundary details into the learning process.

In this study, we evaluated the effectiveness of the proposed methods on three medical image datasets and three segmentation models. Experimental results show that the proposed ATL achieved comparable or superior segmentation performance in many settings compared with conventional Dice Loss and manually tuned Tversky Loss. In particular, on the Synapse multi-organ dataset\cite{synapse} with UNet, ATL improved mIoU by up to 1.83\% compared with Dice Loss. In addition, even when compared with Tversky Loss whose weighting coefficients were individually tuned for each dataset, ATL improved mIoU by 0.13\% without requiring manual tuning. Furthermore, MS-BATL improved mIoU by 2.90\% compared with Dice Loss and by 1.20\% compared with the tuned Tversky Loss. These results suggest that the proposed methods can reduce dataset- and model-dependent bias and contribute to improved generalizability and reliability in medical image segmentation.

The main contributions of this study are summarized as follows.

\begin{itemize}
    \item We propose an adaptive Tversky Loss that dynamically determines the weights for false positives and false negatives based on the Dice coefficient, eliminating the need for manual tuning of the weighting coefficients in Tversky Loss.
    \item We present a loss function design that aims to improve generalizability and reproducibility by mitigating dataset- and model-dependent weight settings.
    \item We introduce a multi-scale boundary-adaptive mechanism that incorporates boundary information as an auxiliary component and investigate its extensibility toward boundary-aware learning.
\end{itemize}

\section{Related Works}
\label{sec:releted}
\subsection{Loss Functions for Class-Imbalanced Segmentation}
In medical image segmentation, the design of loss functions is a crucial factor that significantly affects segmentation performance\cite{segloss}. Cross Entropy Loss is a representative loss function that minimizes pixel-wise classification errors. However, in medical images where background pixels dominate, the learning process may be biased toward the background region, leading to degraded extraction performance for small lesion regions. To address this problem, Dice Loss\cite{dicelosssankou}, which directly evaluates the overlap between predicted and ground-truth regions, has been widely used.

Dice Loss is a loss function based on the Dice coefficient. In multi-class segmentation, the Dice coefficient is computed for each class, and Dice Loss is defined using the average over all classes. In this study, the loss is computed over all classes, including the background class. Dice Loss is formulated as shown in \cref{eq:dice_loss_coefficient}.

\begin{equation}
\label{eq:dice_loss_coefficient}
\small{
\begin{aligned}
L_{\mathrm{Dice}}
&=
1 - \frac{1}{C}\sum_{c=1}^{C} D_{\mathrm{coefficient}}^{(c)},
&
D_{\mathrm{coefficient}}^{(c)}
&=
\frac{2TP^{(c)} + \epsilon}
{2TP^{(c)} + FP^{(c)} + FN^{(c)} + \epsilon}.
\end{aligned}
}
\end{equation}
where $D_{\mathrm{coefficient}}^{(c)}$ in \cref{eq:dice_loss_coefficient} denotes the Dice coefficient for class $c$, $C$ denotes the total number of classes including the background class, and $\epsilon$ is a small constant used to avoid division by zero. In addition, $TP^{(c)}$, $FP^{(c)}$, and $FN^{(c)}$ denote the True Positive, False Positive, and False Negative for class $c$, respectively. These terms are defined using the predicted probability $p_i^{(c)}$ for class $c$ at pixel $i$ and the corresponding one-hot ground-truth label $g_i^{(c)}$ as 

\begin{equation}
\label{eq:tp_fp_fn}
\small{
TP^{(c)} = \sum_{i=1}^{N} p_i^{(c)} g_i^{(c)},
FP^{(c)} = \sum_{i=1}^{N} p_i^{(c)} \left(1 - g_i^{(c)}\right),
FN^{(c)} = \sum_{i=1}^{N} \left(1 - p_i^{(c)}\right) g_i^{(c)}.
}
\end{equation}
where $N$ denotes the number of pixels, $p_i^{(c)} \in [0,1]$ denotes the predicted probability that pixel $i$ belongs to class $c$, and $g_i^{(c)} \in \{0,1\}$ denotes the corresponding ground-truth label. Since Dice Loss directly optimizes the overlap between regions, it is relatively robust to class imbalance and effective for extracting small lesion regions. However, Dice Loss treats $FP$ and $FN$ equally, making it difficult to flexibly reflect task-specific requirements, such as suppressing missed detections or reducing over-segmentation.

To address this limitation, Tversky Loss\cite{tversky} has been proposed as a generalized form of Dice Loss. Tversky Loss assigns different weights to $FP$ and $FN$, thereby controlling the balance between sensitivity and over-segmentation. Multi-class Tversky Loss is formulated by computing the Tversky index for each class and averaging it over all classes, as shown in \cref{eq:tversky_loss}.

\begin{equation}
\label{eq:tversky_loss}
\small{
L_{\mathrm{Tversky}} =
1 -
\frac{1}{C}
\sum_{c=1}^{C}
\frac{TP^{(c)} + \epsilon}
{TP^{(c)} + \alpha FP^{(c)} + \beta FN^{(c)} + \epsilon}.
}
\end{equation}
Where $\alpha$ and $\beta$ are weighting coefficients for $FP^{(c)}$ and $FN^{(c)}$, respectively. Increasing $\beta$ emphasizes the suppression of missed detections, while increasing $\alpha$ emphasizes the suppression of over-segmentation. Therefore, Tversky Loss can flexibly control the type of misclassification according to task-specific characteristics and has the potential to achieve higher performance than Dice Loss\cite{focaltversk}.

However, the performance of Tversky Loss strongly depends on the weighting coefficients $\alpha$ and $\beta$. The optimal settings vary depending on dataset characteristics, target region size, model architecture, and training stage. Therefore, conventional Tversky Loss requires manual tuning of the weighting coefficients for each dataset and model. This manual tuning not only imposes a practical burden but may also lead to settings that are overly adapted to specific data distributions or model architectures. As a result, generalizability to different datasets and models may be degraded.

To address these issues, this study aims to reduce the dependency on manual tuning in Tversky Loss by dynamically determining the weighting coefficients $\alpha$ and $\beta$ based on the Dice coefficient $D_{\mathrm{coefficient}}^{(c)}$ defined in \cref{eq:dice_loss_coefficient}. By doing so, we aim to mitigate bias arising from dataset- and model-dependent weight settings.

\subsection{Boundary-Aware Loss Functions}

In medical image segmentation, accurate boundary estimation is important in addition to region-level overlap. 
This is particularly critical when the boundaries between lesions and normal tissues are ambiguous or when the target region is small, because small boundary errors can substantially affect both evaluation metrics and clinical interpretation. 
Therefore, boundary-aware loss functions have been investigated to complement region-based losses such as Dice Loss and Tversky Loss~\cite{boundary}.
Boundary-aware methods explicitly consider local boundary errors by incorporating information such as boundary distance or contour discrepancies between predicted and ground-truth regions. 
These methods can encourage structural alignment around object boundaries, which is difficult to capture using only region-overlap-based objectives. 
However, boundary information is sensitive to annotation variability, noise, organ shape, lesion size, and imaging conditions. 
Thus, how to effectively and robustly integrate boundary information into loss functions remains an important challenge.
Based on this background, we introduce a multi-scale boundary-adaptive mechanism as an auxiliary extension of ATL. 
Instead of relying on boundary information from a single scale, the proposed MS-BATL incorporates boundary maps extracted using multiple kernel sizes, enabling the model to utilize information ranging from coarse region-level structures to fine boundary details.

\section{Propose Method}
\subsection{Adaptive Tversky Loss}
\label{sec:ATL}
In this section, we describe Adaptive Tversky Loss (ATL), which dynamically adjusts the weighting coefficients of Tversky Loss based on the Dice coefficient. In conventional Tversky Loss, the weighting coefficients $\alpha$ and $\beta$ for false positives (FP) and false negatives (FN) are set to fixed values. However, the optimal weighting coefficients can vary depending on dataset characteristics, target region size, training stage, and model architecture. Therefore, fixed weighting coefficients may not sufficiently adapt to different datasets and models.

To address this issue, we introduce an adaptive weighting strategy that dynamically determines $\alpha$ and $\beta$ based on the class-wise Dice coefficient computed from the current predicted probabilities and the corresponding ground-truth labels within each mini-batch, which represents the current prediction performance during training. Using $D_{\mathrm{coefficient}}^{(c)}$ for class $c$ defined in \cref{eq:dice_loss_coefficient}, the weighting coefficients of ATL are defined as
\begin{equation}
\label{eq:atl_alpha_beta}
\small{
\alpha^{(c)} =
\frac{D_{\mathrm{coefficient}}^{(c)}}{1 + D_{\mathrm{coefficient}}^{(c)}},
\quad
\beta^{(c)} = 1 - \alpha^{(c)}.
}
\end{equation}

\begin{figure}[t]
    \centering
    \includegraphics[width=0.650\linewidth]{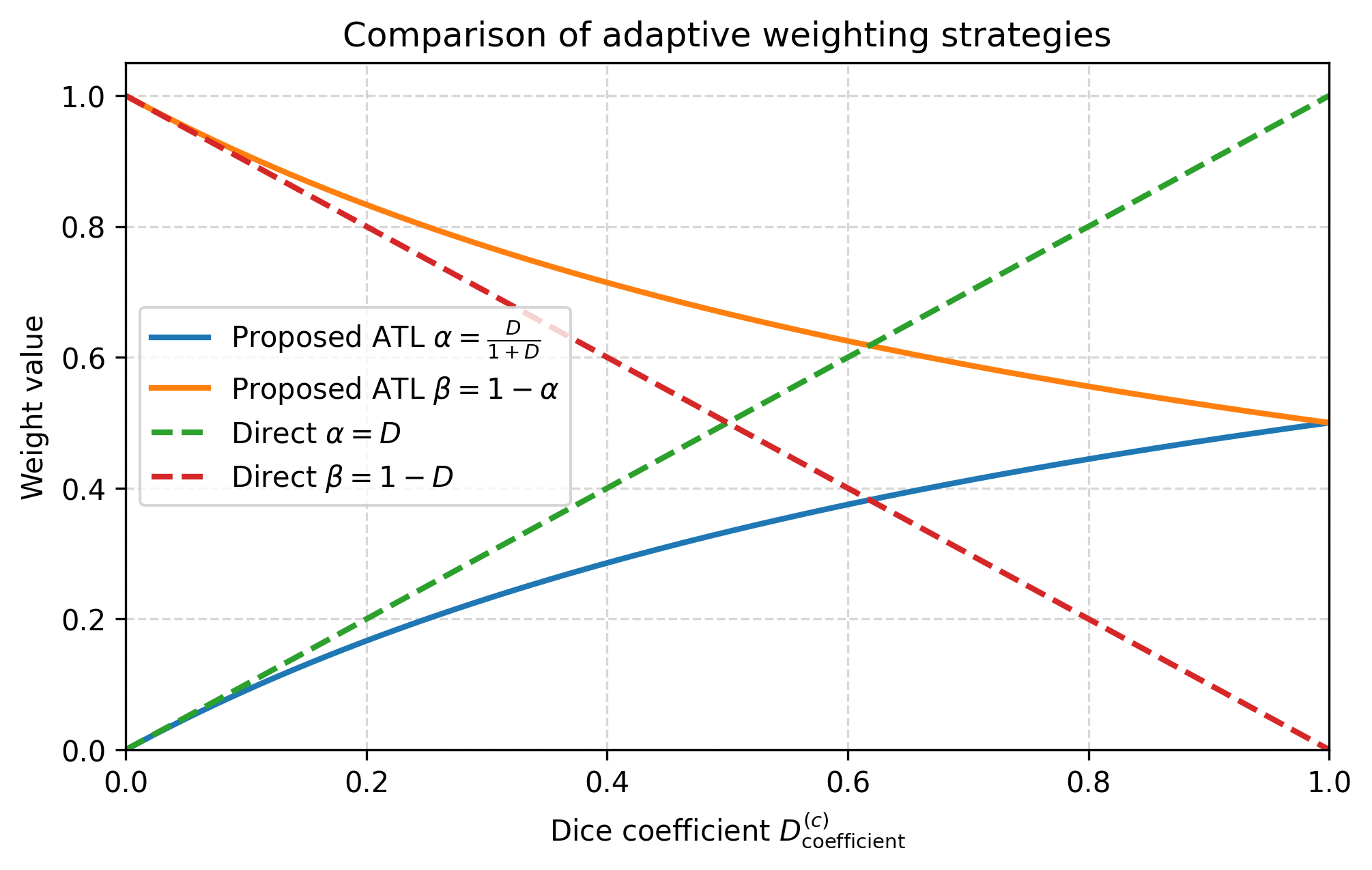}
    \caption{
Relationship between the Dice coefficient and the adaptive weighting coefficients in ATL.
The proposed weighting strategy keeps $\alpha^{(c)} \leq \beta^{(c)}$ for $D_{\mathrm{coefficient}}^{(c)} \in [0,1]$, preventing $\alpha^{(c)}$ from exceeding $\beta^{(c)}$ during training.
This enables a stable transition from FN-oriented optimization to a balanced treatment of FP and FN as prediction performance improves.
For comparison, the direct Dice-based weighting strategy, $\alpha^{(c)}=D_{\mathrm{coefficient}}^{(c)}$, is also shown, where $\alpha^{(c)}$ exceeds $\beta^{(c)}$ once $D_{\mathrm{coefficient}}^{(c)}>0.5$.
}
    \label{fig:atl_weights}
\end{figure}

The relationship between the Dice coefficient and the weighting coefficients $\alpha^{(c)}$ and $\beta^{(c)}$ defined in \cref{eq:atl_alpha_beta} is shown in \cref{fig:atl_weights}.
The proposed nonlinear formulation $\alpha^{(c)} = D_{\mathrm{coefficient}}^{(c)}/(1+D_{\mathrm{coefficient}}^{(c)})$ is designed to keep $\alpha^{(c)} \leq 0.5$ for $D_{\mathrm{coefficient}}^{(c)} \in [0,1]$, while gradually increasing $\alpha^{(c)}$ as prediction performance improves.
In contrast, directly using $\alpha^{(c)} = D_{\mathrm{coefficient}}^{(c)}$ causes $\alpha^{(c)}$ to exceed $\beta^{(c)}$ when the Dice coefficient becomes larger than 0.5, leading to a rapid shift toward FP-oriented weighting.
By preventing $\alpha^{(c)}$ from exceeding $\beta^{(c)}$, ATL maintains FN-oriented optimization in the early training stage and gradually moves toward a balanced treatment of FP and FN.
Furthermore, as $D_{\mathrm{coefficient}}^{(c)}$ approaches 1, both $\alpha^{(c)}$ and $\beta^{(c)}$ approach 0.5, making ATL asymptotically equivalent to Dice Loss.
This is desirable because, when the prediction already has high overlap with the ground truth, the loss should emphasize region overlap rather than excessively prioritizing either FP or FN.

In the early stage of training, the prediction accuracy is low and $D_{\mathrm{coefficient}}^{(c)}$ is also small. As a result, $\alpha^{(c)}$ becomes small and $\beta^{(c)}$ becomes large, which increases the relative penalty for FN and encourages learning to suppress missed target regions. As training progresses and $D_{\mathrm{coefficient}}^{(c)}$ improves, $\alpha^{(c)}$ increases while $\beta^{(c)}$ decreases.

ATL is formulated using the weighting coefficients defined in \cref{eq:atl_alpha_beta} as 
\begin{equation}
\label{eq:atl_loss}
\small{
L_{\mathrm{ATL}}
=
1 -
\frac{1}{C}
\sum_{c=1}^{C}
\frac{TP^{(c)} + \epsilon}
{TP^{(c)} + \alpha^{(c)}FP^{(c)} + \beta^{(c)}FN^{(c)} + \epsilon},
}
\end{equation}
where $C$ denotes the total number of classes including the background class, and $\epsilon$ is a small constant used to avoid division by zero. In addition, $TP^{(c)}$, $FP^{(c)}$, and $FN^{(c)}$ denote the True Positive, False Positive, and False Negative for class $c$, respectively.

In ATL, $\alpha^{(c)}$ and $\beta^{(c)}$ are not treated as independent learnable parameters; instead, they are defined as deterministic functions of the Dice coefficient $D_{\mathrm{coefficient}}^{(c)}$. The Dice coefficient is computed from the current predicted probabilities and the corresponding ground-truth labels, and is therefore not a learnable parameter itself. Since $D_{\mathrm{coefficient}}^{(c)}$ is not detached from the computational graph, gradients can propagate through the Dice coefficient to the network predictions. However, the adaptive weights remain constrained by the functional form in \cref{eq:atl_alpha_beta}, preventing $\alpha^{(c)}$ and $\beta^{(c)}$ from being freely optimized as independent parameters.
From \cref{eq:atl_alpha_beta}, the derivatives of $\alpha^{(c)}$ and $\beta^{(c)}$ with respect to the Dice coefficient are given by
\begin{equation}
\label{eq:atl_alpha_beta_derivative}
\small{
\frac{\partial \alpha^{(c)}}{\partial D_{\mathrm{coefficient}}^{(c)}}
=
\frac{1}{\left(1 + D_{\mathrm{coefficient}}^{(c)}\right)^2},
\quad
\frac{\partial \beta^{(c)}}{\partial D_{\mathrm{coefficient}}^{(c)}}
=
-
\frac{1}{\left(1 + D_{\mathrm{coefficient}}^{(c)}\right)^2}.
}
\end{equation}

The two derivatives have the same magnitude with opposite signs, indicating that $\alpha^{(c)}$ and $\beta^{(c)}$ change in opposite directions while preserving $\alpha^{(c)}+\beta^{(c)}=1$. This property allows ATL to adjust the balance between FP and FN penalties without independently learning the weighting coefficients.

As shown in \cref{eq:atl_alpha_beta_derivative}, the sensitivity of $\alpha^{(c)}$ decreases as $D_{\mathrm{coefficient}}^{(c)}$ increases, which suppresses abrupt changes in the FP/FN weighting balance. As $D_{\mathrm{coefficient}}^{(c)}$ approaches 1, $\alpha^{(c)}$ and $\beta^{(c)}$ both approach 0.5, making ATL asymptotically equivalent to Dice Loss and shifting the optimization toward region-overlap-oriented learning.

If $\alpha$ and $\beta$ were ordinary learnable parameters, they could be optimized in a direction that reduces the loss value but does not necessarily reflect an appropriate FP/FN balance. In contrast, ATL defines $\alpha^{(c)}$ and $\beta^{(c)}$ as deterministic functions of the Dice coefficient, thereby preventing independent over-adaptation of the weighting coefficients. This reduces the need for manual tuning and mitigates dataset- and model-specific weight settings.

\subsection{Multi-scale Boundary-Adaptive Tversky Loss}

In this section, we describe MS-BATL, which extends ATL by incorporating multi-scale boundary information. Since boundary errors can substantially affect medical image segmentation performance, MS-BATL introduces boundary information as an auxiliary component and assigns larger weights to FP and FN around boundary regions.

First, let $G^{(c)}$ denote the ground-truth mask for class $c$. The boundary map $B_k^{(c)}$ for kernel size $k$ is defined based on whether both foreground and background pixels exist within a local neighborhood. 
Specifically, we use a fixed non-negative all-one kernel $\mathbf{K}_k \in \{1\}^{k \times k}$, which is not a learnable convolution kernel. 
\begin{equation}
\label{eq:boundary_map}
\small{
B_k^{(c)}
=
\mathbb{I}
\left[
\mathbf{K}_k * G^{(c)} > 0
\right]
\odot
\mathbb{I}
\left[
\mathbf{K}_k * \left(1-G^{(c)}\right) > 0
\right],
}
\end{equation}

where $*$ denotes convolution, $\mathbb{I}[\cdot]$ is an indicator function, and $\odot$ denotes element-wise multiplication. Since $\mathbf{K}_k$ is a fixed all-one kernel, \cref{eq:boundary_map} extracts pixels whose local neighborhood contains both foreground and background pixels, forming a boundary band around the object contour. We use $k=3,5,7$ to obtain boundary information from fine to coarse scales.

Next, we dynamically determine the scale weights for the boundary maps according to the current prediction performance.
Let $w_3^{(c)}$, $w_5^{(c)}$, and $w_7^{(c)}$ denote the weights for the $3 \times 3$, $5 \times 5$, and $7 \times 7$ boundary maps, respectively.
These weights are designed as triangular functions of the Dice coefficient to reflect the training stage: $w_7^{(c)}$, $w_5^{(c)}$, and $w_3^{(c)}$ correspond to low, intermediate, and high Dice values, respectively.
By using $D_{\mathrm{coefficient}}^{(c)}$, the scale weights are defined as
\begin{equation}
\label{eq:boundary_scale_weights}
\small{
w_3^{(c)} = D_{\mathrm{coefficient}}^{(c)}, \quad
w_7^{(c)} = 1 - D_{\mathrm{coefficient}}^{(c)}, \quad
w_5^{(c)} = 1 - \left|2D_{\mathrm{coefficient}}^{(c)} - 1\right|.
}
\end{equation}
With this formulation, MS-BATL emphasizes coarse boundary information in the early stage of training, mid-scale boundary information during the transition stage, and fine boundary details as prediction performance improves.
The integrated multi-scale boundary map $B^{(c)}$ is then obtained by combining the boundary maps at each scale using normalized weights as
\begin{equation}
\label{eq:multi_scale_boundary_map}
\small{
\begin{aligned}
B^{(c)}
&=
\frac{w_3^{(c)}}{W^{(c)}}B_3^{(c)}
+
\frac{w_5^{(c)}}{W^{(c)}}B_5^{(c)}
+
\frac{w_7^{(c)}}{W^{(c)}}B_7^{(c)},W^{(c)}=w_3^{(c)} + w_5^{(c)} + w_7^{(c)}.
\end{aligned}
}
\end{equation}
where $W^{(c)}$ denotes the sum of the scale weights and normalizes the contribution of each boundary map.
This normalization enables stable integration of multi-scale boundary information without excessively relying on a specific kernel size.

Furthermore, to emphasize boundary regions, we define a pixel-wise boundary weight map $M^{(c)}$ using the multi-scale boundary map $B^{(c)}$ as 
\begin{equation}
\small
\label{eq:boundary_weight_map}
M^{(c)}
=
1
+
B^{(c)}
\left(
1 - D_{\mathrm{coefficient}}^{(c)}
\right).
\end{equation}

In this weight map, larger weights are assigned to pixels around boundary regions where $B^{(c)}$ has high values. By multiplying the boundary map by $1-D_{\mathrm{coefficient}}^{(c)}$, the contribution of boundary regions becomes relatively larger in the early stage of training, when prediction performance is low. In contrast, as training progresses and $D_{\mathrm{coefficient}}^{(c)}$ becomes higher, the influence of the boundary weights decreases, suppressing excessive dependence on boundary information.

In MS-BATL, FP and FN are recalculated using the boundary weight map $M^{(c)}$. The boundary-weighted FP and FN are defined as 
\begin{equation}
\label{eq:boundary_weighted_fp_fn}
\small{
\begin{aligned}
FP_b^{(c)}&=\sum_{i=1}^{N}M_i^{(c)}p_i^{(c)}\left(1 - g_i^{(c)}\right), FN_b^{(c)}=\sum_{i=1}^{N}M_i^{(c)}\left(1 - p_i^{(c)}\right)g_i^{(c)}.
\end{aligned}
}
\end{equation}
where $M_i^{(c)}$ denotes the boundary weight for class $c$ at pixel $i$, $p_i^{(c)}$ denotes the predicted probability that pixel $i$ belongs to class $c$, and $g_i^{(c)}$ denotes the corresponding one-hot ground-truth label. The boundary weights are applied to FP and FN, but not to TP, because the purpose is to place greater emphasis on misclassifications around boundary regions.

Finally, MS-BATL is defined using the adaptive weighting coefficients $\alpha^{(c)}$ and $\beta^{(c)}$ introduced in ATL as
\begin{equation}
\label{eq:ms_batl_loss}
\small{
L_{\mathrm{MS\mbox{-}BATL}}
=
1
-
\frac{1}{C}
\sum_{c=1}^{C}
\frac{
TP^{(c)} + \epsilon
}
{
TP^{(c)}
+
\alpha^{(c)}FP_b^{(c)}
+
\beta^{(c)}FN_b^{(c)}
+
\epsilon
}.
}
\end{equation}
where $\alpha^{(c)}$ and $\beta^{(c)}$ are the adaptive weighting coefficients of ATL defined in \cref{eq:atl_alpha_beta}. These coefficients are determined based on the Dice coefficient $D_{\mathrm{coefficient}}^{(c)}$ computed from $TP^{(c)}$, $FP^{(c)}$, and $FN^{(c)}$ without applying boundary weights. This allows MS-BATL to preserve the dynamic control of FP and FN provided by ATL while additionally emphasizing errors around boundary regions. As a result, MS-BATL enables learning that considers boundary information in addition to region-overlap-based optimization.

\section{Experiments}
\label{sec:experiments}
\subsection{Experimental Setup}
In this study, we used the Kvasir-SEG dataset \cite{kvasir} for binary segmentation, and the ACDC dataset \cite{ACDC} and the Synapse multi-organ dataset \cite{synapse} for multi-class segmentation.
Kvasir-SEG consists of 1,000 endoscopic polyp images and was split into 80\% for training, 10\% for validation, and 10\% for testing. 

ACDC contains cardiac MRI images with annotations for the left ventricle, right ventricle, and myocardium, and we used 70 cases for training, 10 for validation, and 20 for testing. 
The Synapse dataset consists of 30 abdominal CT volumes, and eight organs were evaluated. The dataset was divided into 18 cases for training, 6 for validation, and 6 for testing.
For ACDC and Synapse, 5-fold cross-validation was performed, and the mean and standard deviation across folds were reported. 

We compared Dice Loss \cite{dicelosssankou}, Tversky Loss \cite{tversky}, Noise-robust Dice Loss \cite{noiserobustdice}, Focal Tversky Loss \cite{focaltversk}, and Boundary Loss + Dice Loss \cite{boundary}. 
In addition, we applied the proposed ATL and MS-BATL to Noise-robust Dice Loss and Focal Tversky Loss on the Synapse dataset to examine their extensibility.

We used U-Net \cite{Unet}, TransUNet \cite{transunet}, and Swin-UNet \cite{swinunet} as segmentation models. All models were trained from scratch, and the weights with the highest validation mIoU were used for testing.
Intersection over Union (IoU) was used as the evaluation metric, and mean IoU (mIoU) was used as the primary metric.

The batch size was set to 24, and training was performed for 4,000 iterations using SGD with a momentum of 0.9 and a weight decay of $2 \times 10^{-4}$. 
The initial learning rate was set to $0.01$ and decayed using a polynomial schedule $lr_{\mathrm{iter}} = 0.01 \left(1-\mathrm{iter}/\mathrm{iter}_{\max}\right)^{0.9}$.
All images were resized to $224 \times 224$, normalized to [0, 1], and augmented using horizontal flipping and random rotation in $[-90^\circ, 90^\circ]$. 
All experiments were conducted on a single NVIDIA RTX A6000 GPU.

\begin{table*}[t]
\centering
\caption{
Segmentation performance (IoU, \%) of different loss functions on the Kvasir-SEG and ACDC datasets.
For Kvasir-SEG, only mIoU is reported because it is a binary segmentation dataset.
For ACDC, mIoU and class-wise IoU are reported.
}
\label{tab:kvasir_acdc}
\tiny
\resizebox{\textwidth}{!}{
\begin{tabular}{llcccccc}
\toprule
\multirow{2}{*}{Networks}
&
\multirow{2}{*}{Loss function}
&
\multicolumn{1}{c}{Kvasir-SEG}
&
\multicolumn{5}{c}{ACDC}\\
\cmidrule(lr){3-3}\cmidrule(lr){4-8}
&&mIoU&mIoU&background&Right ventricle&Myocardium&Left ventricle\\
\midrule

U\text{-}Net 
& Dice Loss 
& 83.60
& 86.47$\pm$2.33 & 99.56$\pm$0.10 & 82.94$\pm$4.18 & 73.60$\pm$2.37 & 89.78$\pm$3.42 \\

& Tversky Loss ($\alpha$=0.3,$\beta$=1-$\alpha$)
& 82.60
& 86.73$\pm$2.12 & 99.56$\pm$0.11 & 82.72$\pm$4.38 & 74.20$\pm$2.60 & 90.43$\pm$2.50 \\

& Tversky Loss ($\alpha$=0.7,$\beta$=1-$\alpha$)
& 81.62
& 86.32$\pm$2.39 & 99.53$\pm$0.11 & 81.82$\pm$5.03 & 73.28$\pm$2.81 & \textcolor{red}{90.65$\pm$2.75} \\

& Focal Tversky Loss ($\gamma$=2,$\alpha$=0.3,$\beta$=1-$\alpha$)
& 77.96
& 84.04$\pm$2.63 & 99.42$\pm$0.14 & 78.65$\pm$5.91 & 69.61$\pm$3.16 & 88.49$\pm$2.87 \\

& Focal Tversky Loss ($\gamma$=2,$\alpha$=0.5,$\beta$=1-$\alpha$)
& 77.93
& 84.54$\pm$2.49 & 99.46$\pm$0.12 & 79.14$\pm$4.87 & 70.63$\pm$3.04 & 88.93$\pm$2.72 \\

& Focal Tversky Loss ($\gamma$=2,$\alpha$=0.7,$\beta$=1-$\alpha$)
& 77.95
& 83.90$\pm$2.98 & 99.42$\pm$0.14 & 77.94$\pm$5.32 & 69.50$\pm$3.57 & 88.75$\pm$3.64 \\

& Boundary Loss + Dice Loss
& \textcolor{red}{83.88}
& 86.47$\pm$2.29 & 99.56$\pm$0.10 & 82.90$\pm$4.33 & 73.61$\pm$2.41 & 89.80$\pm$3.09 \\

& Noise-robust Dice Loss
& 82.53
& 86.64$\pm$2.59 & 99.55$\pm$0.12 & 82.61$\pm$5.10 & 74.09$\pm$2.76 & 90.31$\pm$3.01 \\

& ATL (Ours)
& 83.45
& 86.76$\pm$2.36 & 99.56$\pm$0.10 & \textcolor{red}{83.11$\pm$4.76} & 74.14$\pm$2.29 & 90.23$\pm$3.05 \\

& MS-BATL (Ours)
& 83.45
& \textcolor{red}{86.83$\pm$2.25} & 99.56$\pm$0.11 & 82.94$\pm$4.58 & \textcolor{red}{74.52$\pm$2.56} & 90.30$\pm$2.76 \\

\midrule

TransUNet
& Dice Loss
& 82.20
& 85.19$\pm$3.07 & 99.49$\pm$0.15 & 77.53$\pm$8.80 & \textcolor{red}{73.61$\pm$1.60} & \textcolor{red}{90.13$\pm$2.28} \\

& Tversky Loss ($\alpha$=0.3,$\beta$=1-$\alpha$)
& 81.80
& 85.29$\pm$2.37 & 99.46$\pm$0.14 & 80.70$\pm$5.96 & 71.67$\pm$2.82 & 89.34$\pm$2.11 \\

& Tversky Loss ($\alpha$=0.7,$\beta$=1-$\alpha$)
& 81.97
& 84.57$\pm$2.58 & 99.41$\pm$0.17 & 79.80$\pm$4.73 & 70.03$\pm$4.27 & 89.03$\pm$2.27 \\

& Focal Tversky Loss ($\gamma$=2,$\alpha$=0.3,$\beta$=1-$\alpha$)
& 79.47
& 82.58$\pm$2.89 & 99.33$\pm$0.16 & 76.16$\pm$7.09 & 67.83$\pm$3.19 & 86.99$\pm$3.29 \\

& Focal Tversky Loss ($\gamma$=2,$\alpha$=0.5,$\beta$=1-$\alpha$)
& 78.82
& 83.19$\pm$3.24 & 99.37$\pm$0.19 & 77.40$\pm$7.87 & 68.65$\pm$2.76 & 87.35$\pm$3.08 \\

& Focal Tversky Loss ($\gamma$=2,$\alpha$=0.7,$\beta$=1-$\alpha$)
& 77.94
& 82.48$\pm$3.27 & 99.32$\pm$0.21 & 76.41$\pm$6.74 & 67.53$\pm$3.69 & 86.65$\pm$3.26 \\

& Boundary Loss + Dice Loss
& 83.52
& 85.44$\pm$2.65 & 99.47$\pm$0.19 & 80.29$\pm$5.81 & 72.35$\pm$3.43 & 89.66$\pm$2.33 \\

& Noise-robust Dice Loss
& 83.10
& \textcolor{red}{86.04$\pm$2.65} & \textcolor{red}{99.50$\pm$0.16} & \textcolor{red}{81.70$\pm$6.83} & 72.88$\pm$2.56 & 90.07$\pm$2.26 \\

& ATL (Ours)
& 82.56
& 85.72$\pm$2.48 & 99.48$\pm$0.17 & 81.57$\pm$5.72 & 72.12$\pm$3.37 & 89.72$\pm$2.05 \\

& MS-BATL (Ours)
& \textcolor{red}{83.56}
& 85.54$\pm$2.34 & 99.49$\pm$0.16 & 80.87$\pm$5.71 & 72.39$\pm$2.98 & 89.41$\pm$2.25 \\

\midrule

Swin-UNet
& Dice Loss
& 65.86
& 71.35$\pm$3.79 & 98.73$\pm$0.23 & 59.90$\pm$10.02 & \textcolor{red}{51.58$\pm$2.83} & 75.17$\pm$3.85 \\

& Tversky Loss ($\alpha$=0.3,$\beta$=1-$\alpha$)
& 66.15
& 70.12$\pm$2.78 & 98.62$\pm$0.17 & 57.32$\pm$8.50 & 50.76$\pm$2.86 & 73.78$\pm$3.38 \\

& Tversky Loss ($\alpha$=0.7,$\beta$=1-$\alpha$)
& 64.73
& 70.52$\pm$4.19 & 98.71$\pm$0.29 & 58.94$\pm$9.26 & 49.51$\pm$4.59 & 74.91$\pm$3.95 \\

& Focal Tversky Loss ($\gamma$=2,$\alpha$=0.3,$\beta$=1-$\alpha$)
& 65.00
& 67.16$\pm$2.72 & 98.37$\pm$0.17 & 53.65$\pm$9.63 & 46.80$\pm$3.76 & 69.83$\pm$3.57 \\

& Focal Tversky Loss ($\gamma$=2,$\alpha$=0.5,$\beta$=1-$\alpha$)
& 65.46
& 68.15$\pm$3.81 & 98.54$\pm$0.26 & 54.91$\pm$10.57 & 48.19$\pm$3.79 & 70.95$\pm$3.60 \\

& Focal Tversky Loss ($\gamma$=2,$\alpha$=0.7,$\beta$=1-$\alpha$)
& 64.73
& 68.23$\pm$3.46 & 98.59$\pm$0.26 & 55.54$\pm$8.58 & 47.06$\pm$3.29 & 71.72$\pm$3.79 \\

& Boundary Loss + Dice Loss
& 64.79
& 71.28$\pm$3.50 & 98.76$\pm$0.21 & 60.24$\pm$8.09 & 51.23$\pm$3.22 & 74.86$\pm$4.41 \\

& Noise-robust Dice Loss
& 65.37
& \textcolor{red}{71.69$\pm$3.43} & \textcolor{red}{98.79$\pm$0.20} & \textcolor{red}{61.13$\pm$8.62} & 51.51$\pm$3.14 & \textcolor{red}{75.33$\pm$4.10} \\

& ATL (Ours)
& \textcolor{red}{66.30}
& 70.51$\pm$4.06 & 98.67$\pm$0.26 & 58.70$\pm$10.89 & 50.52$\pm$3.39 & 74.14$\pm$4.22 \\

& MS-BATL (Ours)
& 66.29
& 70.87$\pm$3.73 & 98.69$\pm$0.25 & 59.33$\pm$10.61 & 51.13$\pm$2.98 & 74.34$\pm$4.31 \\

\bottomrule
\end{tabular}
}
\end{table*}

\begin{table}[t]
\centering
\caption{Segmentation performance (IoU, \%) for different class and loss functions on the  Synapse multi-organ dataset.}
\label{tab:synapse}
\resizebox{\textwidth}{!}{%
\begin{tabular}{lccccccccccccc}
\toprule
Networks & Loss function & mIoU  & background & Aorta  & Gallbladder  & Kidney(L) & Kiedney(R) & Liver & Pancreas & Spleen & Stomach \\
\midrule

U\text{-}Net & Dice Loss &58.11$\pm$8.44&98.84$\pm$0.26&66.18$\pm$7.04&42.84$\pm$14.88&49.86$\pm$10.87&47.97$\pm$10.77&75.99$\pm$10.18&32.04$\pm$7.50&61.46$\pm$24.28&47.84$\pm$5.45\\
& Tversky Loss($\alpha$=0.3,$\beta$=1-$\alpha$) &59.81$\pm$8.60&98.81$\pm$0.23&66.46$\pm$6.72&\textcolor{red}{45.03$\pm$12.94}&55.55$\pm$12.14&50.78$\pm$15.13&75.67$\pm$8.56&35.57$\pm$4.04&62.17$\pm$24.34&48.25$\pm$6.01\\
& Tversky Loss($\alpha$=0.7,$\beta$=1-$\alpha$) &56.07$\pm$7.81&98.72$\pm$0.27&63.06$\pm$10.68&42.09$\pm$13.72&51.16$\pm$9.83&46.47$\pm$10.89&74.80$\pm$9.74&26.20$\pm$7.45&58.41$\pm$22.96&43.72$\pm$4.75\\
& Focal Tversky Loss($\gamma$=2,$\alpha$=0.3,$\beta$=1-$\alpha$) &57.24$\pm$8.57&98.72$\pm$0.25&64.67$\pm$8.93&39.13$\pm$14.03&52.64$\pm$10.43&46.55$\pm$13.64&73.08$\pm$9.69&33.71$\pm$4.82&60.33$\pm$24.36&46.34$\pm$7.26\\
& Focal Tversky Loss($\gamma$=2,$\alpha$=0.5,$\beta$=1-$\alpha$) &55.37$\pm$7.92&98.69$\pm$0.31&61.78$\pm$10.80&37.84$\pm$12.65&47.95$\pm$10.01&41.99$\pm$12.39&73.40$\pm$10.09&32.69$\pm$4.70&59.49$\pm$24.32&44.50$\pm$8.18\\
& Focal Tversky Loss($\gamma$=2,$\alpha$=0.7,$\beta$=1-$\alpha$) &52.51$\pm$7.33&98.50$\pm$0.33&57.89$\pm$11.23&38.78$\pm$12.45&48.21$\pm$7.76&38.42$\pm$14.94&69.56$\pm$10.30&25.78$\pm$5.33&56.97$\pm$24.47&38.51$\pm$10.16\\
& Boundary Loss + Dice Loss &57.31$\pm$8.34&98.78$\pm$0.40&64.37$\pm$9.40&38.22$\pm$10.38&50.41$\pm$14.21&46.91$\pm$14.81&75.17$\pm$12.45&32.71$\pm$5.35&62.03$\pm$23.88&47.21$\pm$8.40\\
& Noise robust Dice Loss &56.81$\pm$7.92&98.73$\pm$0.31&62.31$\pm$9.87&39.37$\pm$13.84&51.76$\pm$7.11&46.54$\pm$13.79&74.72$\pm$10.45&31.96$\pm$3.88&61.34$\pm$24.11&44.57$\pm$8.04\\
& Noise robust Dice Loss(ATL$_{ver}$) (Ours) &59.47$\pm$8.98&98.82$\pm$0.40&64.96$\pm$9.31&43.96$\pm$14.86&56.07$\pm$11.63&51.93$\pm$15.93&76.84$\pm$10.86&31.97$\pm$6.88&62.52$\pm$23.18&48.17$\pm$8.98\\
& Noise robust Dice Loss(MS-BATL$_{ver}$) (Ours) &58.13$\pm$8.29&98.79$\pm$0.24&65.19$\pm$8.81&44.41$\pm$15.40&53.57$\pm$12.61&50.44$\pm$12.46&75.93$\pm$8.33&28.19$\pm$6.40&61.13$\pm$22.10&45.51$\pm$8.08\\
& Focal Tversky Loss(ATL$_{ver}$) (Ours) &57.22$\pm$6.33&98.74$\pm$0.27&63.76$\pm$4.50&42.01$\pm$9.11&53.16$\pm$9.62&46.02$\pm$13.88&73.54$\pm$9.64&31.87$\pm$5.02&60.89$\pm$23.78&44.98$\pm$8.56\\
& Focal Tversky Loss(MS-BATL$_{ver}$) (Ours) &53.52$\pm$6.98&98.74$\pm$0.25&62.64$\pm$9.88&9.80$\pm$21.91&52.82$\pm$13.41&46.73$\pm$15.42&72.94$\pm$7.32&31.79$\pm$5.25&58.87$\pm$24.72&47.32$\pm$8.91\\
& ATL(Ours) &59.94$\pm$7.96&98.85$\pm$0.32&\textcolor{red}{66.97$\pm$7.42}&44.76$\pm$10.45&53.30$\pm$13.18&50.37$\pm$15.61&76.64$\pm$11.41&33.86$\pm$5.84&66.36$\pm$26.37&48.32$\pm$7.49\\
& MS-BATL(Ours) &\textcolor{red}{61.01$\pm$9.59}&\textcolor{red}{98.87$\pm$0.24}&65.17$\pm$9.95&39.52$\pm$16.83&\textcolor{red}{60.97$\pm$13.77}&\textcolor{red}{57.44$\pm$16.03}&\textcolor{red}{77.56$\pm$8.96}&\textcolor{red}{35.31$\pm$5.38}&\textcolor{red}{64.39$\pm$25.64}&\textcolor{red}{49.86$\pm$7.60}\\
\midrule

TransUNet & Dice Loss &60.26$\pm$6.80&98.90$\pm$0.06&66.18$\pm$7.62&41.08$\pm$7.72&56.09$\pm$9.27&58.68$\pm$12.60&77.36$\pm$8.09&30.87$\pm$2.74&65.40$\pm$24.26&47.79$\pm$4.29\\
& Tversky Loss($\alpha$=0.3,$\beta$=1-$\alpha$) &61.03$\pm$7.80&98.90$\pm$0.05&65.78$\pm$6.56&42.47$\pm$10.92&59.16$\pm$9.42&60.71$\pm$11.62&77.97$\pm$8.43&32.50$\pm$2.37&63.85$\pm$27.78&47.93$\pm$3.51\\
& Tversky Loss($\alpha$=0.7,$\beta$=1-$\alpha$) &58.49$\pm$7.62&98.77$\pm$0.17&61.88$\pm$7.59&44.57$\pm$11.96&55.78$\pm$9.19&54.67$\pm$13.26&75.25$\pm$8.96&28.18$\pm$6.83&62.27$\pm$25.86&45.05$\pm$3.38\\
& Focal Tversky Loss($\gamma$=2,$\alpha$=0.3,$\beta$=1-$\alpha$) &56.43$\pm$7.73&98.60$\pm$0.14&60.13$\pm$8.24&40.62$\pm$12.92&52.84$\pm$10.78&51.17$\pm$12.33&72.76$\pm$8.16&29.24$\pm$4.54&58.02$\pm$23.41&44.46$\pm$4.33\\
& Focal Tversky Loss($\gamma$=2,$\alpha$=0.5,$\beta$=1-$\alpha$) &57.49$\pm$7.58&98.67$\pm$0.16&60.80$\pm$6.56&38.56$\pm$13.01&55.53$\pm$8.29&54.10$\pm$10.25&74.24$\pm$8.47&30.09$\pm$4.70&61.57$\pm$24.16&43.89$\pm$3.50\\
& Focal Tversky Loss($\gamma$=2,$\alpha$=0.7,$\beta$=1-$\alpha$) &56.40$\pm$7.05&98.57$\pm$0.16&56.84$\pm$7.60&37.44$\pm$11.60&56.19$\pm$9.33&55.89$\pm$14.00&70.47$\pm$7.55&29.36$\pm$4.24&60.35$\pm$23.24&42.53$\pm$2.50\\
& Boundary Loss + Dice Loss &60.50$\pm$6.16&98.86$\pm$0.22&64.23$\pm$7.14&41.87$\pm$6.12&59.46$\pm$9.39&58.20$\pm$10.96&76.56$\pm$9.90&31.31$\pm$3.56&66.09$\pm$24.23&47.89$\pm$2.96\\
& Noise robust Dice Loss &60.07$\pm$8.23&98.86$\pm$0.10&62.64$\pm$7.52&38.28$\pm$15.78&61.27$\pm$10.07&59.97$\pm$14.25&77.32$\pm$10.03&32.59$\pm$3.23&63.55$\pm$25.84&46.14$\pm$3.18\\
& Noise robust Dice Loss(ATL$_{ver}$) (Ours) &61.30$\pm$7.09&98.91$\pm$0.10&66.83$\pm$5.86&43.34$\pm$12.18&57.82$\pm$8.35&59.24$\pm$11.95&78.38$\pm$8.22&32.32$\pm$3.01&66.87$\pm$25.28&48.02$\pm$4.16\\
& Noise robust Dice Loss(MS-BATL$_{ver}$) (Ours) &62.21$\pm$7.56&98.91$\pm$0.09&64.75$\pm$7.61&43.01$\pm$15.69&\textcolor{red}{61.68$\pm$7.75}&61.74$\pm$9.82&\textcolor{red}{79.15$\pm$7.69}&34.22$\pm$2.76&\textcolor{red}{68.19$\pm$26.47}&48.24$\pm$4.14\\
& Focal Tversky Loss(ATL$_{ver}$) (Ours) &58.16$\pm$7.25&98.73$\pm$0.13&61.55$\pm$7.67&36.81$\pm$10.35&56.61$\pm$10.52&53.79$\pm$13.47&75.11$\pm$8.70&31.77$\pm$4.08&62.65$\pm$24.85&46.40$\pm$5.40\\
& Focal Tversky Loss(MS-BATL$_{ver}$) (Ours) &49.35$\pm$8.71&98.64$\pm$0.18&53.64$\pm$30.11&0.00$\pm$0.00&56.62$\pm$8.20&55.43$\pm$9.96&72.54$\pm$10.83&0.00$\pm$0.00&62.80$\pm$26.80&44.49$\pm$3.64\\
& ATL(Ours) &61.35$\pm$7.44&98.90$\pm$0.12&65.69$\pm$6.51&43.44$\pm$9.66&59.14$\pm$6.69&59.97$\pm$9.83&77.14$\pm$8.52&32.95$\pm$6.05&66.43$\pm$26.28&48.49$\pm$4.84\\
& MS-BATL(Ours) &\textcolor{red}{62.59$\pm$8.44}&\textcolor{red}{98.95$\pm$0.11}&\textcolor{red}{67.55$\pm$6.92}&\textcolor{red}{46.47$\pm$11.71}&59.90$\pm$11.25&\textcolor{red}{61.91$\pm$13.00}&78.58$\pm$8.40&\textcolor{red}{35.01$\pm$3.38}&65.57$\pm$27.96&\textcolor{red}{49.33$\pm$4.75}\\

\midrule

Swin-UNet & Dice Loss &\textcolor{red}{42.20$\pm$7.01}&\textcolor{red}{97.84$\pm$0.21}&33.49$\pm$6.92&29.04$\pm$13.25&\textcolor{red}{32.86$\pm$12.83}&\textcolor{red}{33.88$\pm$13.59}&\textcolor{red}{61.95$\pm$7.65}&17.59$\pm$4.21&42.85$\pm$17.56&30.33$\pm$3.23\\
& Tversky Loss($\alpha$=0.3,$\beta$=1-$\alpha$) &41.18$\pm$6.59&97.46$\pm$0.43&33.06$\pm$7.15&28.07$\pm$14.59&31.89$\pm$10.81&32.36$\pm$10.78&60.75$\pm$7.55&16.88$\pm$3.68&41.45$\pm$15.23&28.74$\pm$3.79\\
& Tversky Loss($\alpha$=0.7,$\beta$=1-$\alpha$) &40.60$\pm$6.29&97.83$\pm$0.22&31.98$\pm$6.43&28.06$\pm$12.48&29.83$\pm$12.44&30.68$\pm$10.50&59.90$\pm$7.05&16.43$\pm$3.70&41.22$\pm$15.88&29.50$\pm$4.78\\
& Focal Tversky Loss($\gamma$=2,$\alpha$=0.3,$\beta$=1-$\alpha$) &40.20$\pm$6.15&97.23$\pm$0.40&32.84$\pm$7.73&30.26$\pm$12.70&29.16$\pm$11.02&29.78$\pm$10.75&57.79$\pm$7.38&17.80$\pm$2.98&38.80$\pm$14.01&28.15$\pm$3.81\\
& Focal Tversky Loss($\gamma$=2,$\alpha$=0.5,$\beta$=1-$\alpha$) &41.87$\pm$6.72&97.67$\pm$0.27&\textcolor{red}{33.82$\pm$8.34}&\textcolor{red}{32.17$\pm$11.98}&31.30$\pm$12.63&32.45$\pm$12.53&59.66$\pm$6.97&\textcolor{red}{18.52$\pm$2.97}&41.09$\pm$15.59&30.18$\pm$3.29\\
& Focal Tversky Loss($\gamma$=2,$\alpha$=0.7,$\beta$=1-$\alpha$) &40.22$\pm$7.08&97.17$\pm$1.51&32.01$\pm$7.71&22.26$\pm$16.64&30.41$\pm$12.30&33.07$\pm$13.34&57.69$\pm$9.19&17.75$\pm$3.62&41.60$\pm$17.04&30.06$\pm$5.05\\
& Boundary Loss + Dice Loss &42.17$\pm$6.92&97.93$\pm$0.15&33.44$\pm$6.79&29.00$\pm$12.33&32.38$\pm$12.36&33.52$\pm$13.45&62.02$\pm$7.43&17.77$\pm$4.02&\textcolor{red}{43.11$\pm$17.60}&\textcolor{red}{30.40$\pm$3.32}\\
& Noise robust Dice Loss &39.73$\pm$5.71&97.72$\pm$0.12&32.62$\pm$7.07&22.15$\pm$8.52&27.07$\pm$10.94&30.53$\pm$11.51&59.78$\pm$7.08&16.99$\pm$2.87&42.01$\pm$17.27&28.73$\pm$3.78\\
& Noise robust Dice Loss(ATL$_{ver}$) (Ours) &33.97$\pm$5.20&88.99$\pm$2.27&29.81$\pm$4.01&0.38$\pm$0.15&25.86$\pm$8.19&21.06$\pm$16.29&58.29$\pm$7.36&14.88$\pm$1.95&39.55$\pm$15.43&26.90$\pm$5.10\\
& Noise robust Dice Loss(MS-BATL$_{ver}$) (Ours) &31.54$\pm$4.26&91.23$\pm$2.28&26.21$\pm$4.92&0.62$\pm$0.28&16.23$\pm$8.78&21.36$\pm$12.74&56.64$\pm$7.77&13.28$\pm$2.80&37.64$\pm$12.57&20.66$\pm$12.89\\
& Focal Tversky Loss(ATL$_{ver}$) (Ours) &35.92$\pm$4.38&89.65$\pm$5.68&31.65$\pm$5.56&0.54$\pm$0.35&28.83$\pm$9.04&29.98$\pm$9.82&56.02$\pm$7.03&17.21$\pm$2.70&41.52$\pm$16.47&27.85$\pm$3.86\\
& Focal Tversky Loss(MS-BATL$_{ver}$) (Ours) &37.04$\pm$6.02&96.71$\pm$1.22&25.92$\pm$7.00&17.39$\pm$11.98&30.10$\pm$10.30&29.52$\pm$6.56&57.75$\pm$8.32&8.11$\pm$7.54&41.64$\pm$16.98&26.24$\pm$2.39\\
& ATL(Ours) &37.93$\pm$4.90&96.92$\pm$0.69&31.97$\pm$4.20&3.36$\pm$2.34&29.05$\pm$9.24&31.30$\pm$9.95&60.79$\pm$7.59&16.01$\pm$2.32&42.81$\pm$15.38&29.20$\pm$5.01\\
& MS-BATL(Ours) &36.27$\pm$3.75&95.54$\pm$1.17&30.85$\pm$3.47&1.88$\pm$1.33&28.91$\pm$7.71&22.81$\pm$11.96&59.74$\pm$6.82&14.95$\pm$2.30&42.81$\pm$14.96&28.95$\pm$4.66\\
\midrule
\bottomrule
\end{tabular}}
\end{table}

\subsection{Quantitative Evaluation}
\cref{tab:kvasir_acdc} presents the quantitative evaluation results on the Kvasir-SEG and ACDC datasets.
On Kvasir-SEG, ATL and MS-BATL achieved results that were 0.15\% lower than Dice Loss with U-Net. In contrast, MS-BATL improved the performance by 1.36\% with TransUNet, and ATL achieved the best performance with Swin-UNet, reaching an mIoU of 66.30\%. On ACDC, MS-BATL achieved an mIoU of 86.83\% with U-Net, outperforming both Dice Loss and fixed-weight Tversky Loss. In addition, ATL achieved the best IoU for the right ventricle, while MS-BATL achieved the best IoU for the myocardium. These results confirm the effectiveness of adaptive weighting according to class-specific misclassification tendencies.

In contrast, for TransUNet and Swin-UNet on ACDC, Noise-robust Dice Loss achieved the best performance, and ATL and MS-BATL did not always achieve the highest accuracy. However, compared with Dice Loss and fixed-weight Tversky Loss, ATL and MS-BATL achieved comparable or better performance in many settings, suggesting that they can serve as effective alternatives without manual tuning. In particular, performance degradation was observed with Swin-UNet, indicating that the proposed methods tend to depend on prediction stability and model architecture, because the weights are determined based on the Dice coefficient.

\cref{tab:synapse} shows the results on the Synapse multi-organ dataset.
ATL and MS-BATL outperformed Dice Loss by 1.83\% and 2.90\% with U-Net, and MS-BATL achieved the best performance with an mIoU of 61.01\%. With TransUNet, MS-BATL also achieved 62.59\%, outperforming the other methods. These results demonstrate that the dynamic weighting of ATL and the incorporation of boundary information in MS-BATL are effective for multi-organ segmentation.

On the other hand, on Synapse with Swin-UNet, all loss functions showed relatively low performance compared with U-Net and TransUNet. Although Dice Loss achieved the highest mIoU in this setting, this may be partly because Swin-UNet did not fully benefit from its representation capacity when trained from scratch. We trained all models from scratch to fairly evaluate the intrinsic effect of each loss function without the influence of pretrained weights. Under unstable prediction conditions, especially for small organs, the Dice-coefficient-based adaptive weighting in ATL and the boundary weighting in MS-BATL may become less reliable, limiting their effectiveness.

Furthermore, when the ATL and MS-BATL frameworks were applied to Noise-robust Dice Loss, performance improvements were observed with U-Net and TransUNet. In particular, with TransUNet, the MS-BATL variant achieved 62.21\%, compared with 60.07\% for the original Noise-robust Dice Loss. This result suggests that the proposed adaptive weighting mechanism can be extended to existing Dice-based loss functions. In contrast, applying MS-BATL to Focal Tversky Loss led to performance degradation for some small organs. Since Focal Tversky Loss includes exponential modulation, its interaction with Dice-coefficient-based dynamic weighting and boundary weighting may have caused instability in the loss scale or gradients.

Overall, ATL provides stable performance without parameter tuning, while MS-BATL can bring substantial improvements when boundary information is effective. However, careful consideration is required regarding the conditions under which MS-BATL is applied.

\subsection{Qualitative evaluation}

\begin{figure*}[t]
    \centering

    \begin{minipage}[t]{0.48\textwidth}
        \centering
        \includegraphics[width=\linewidth]{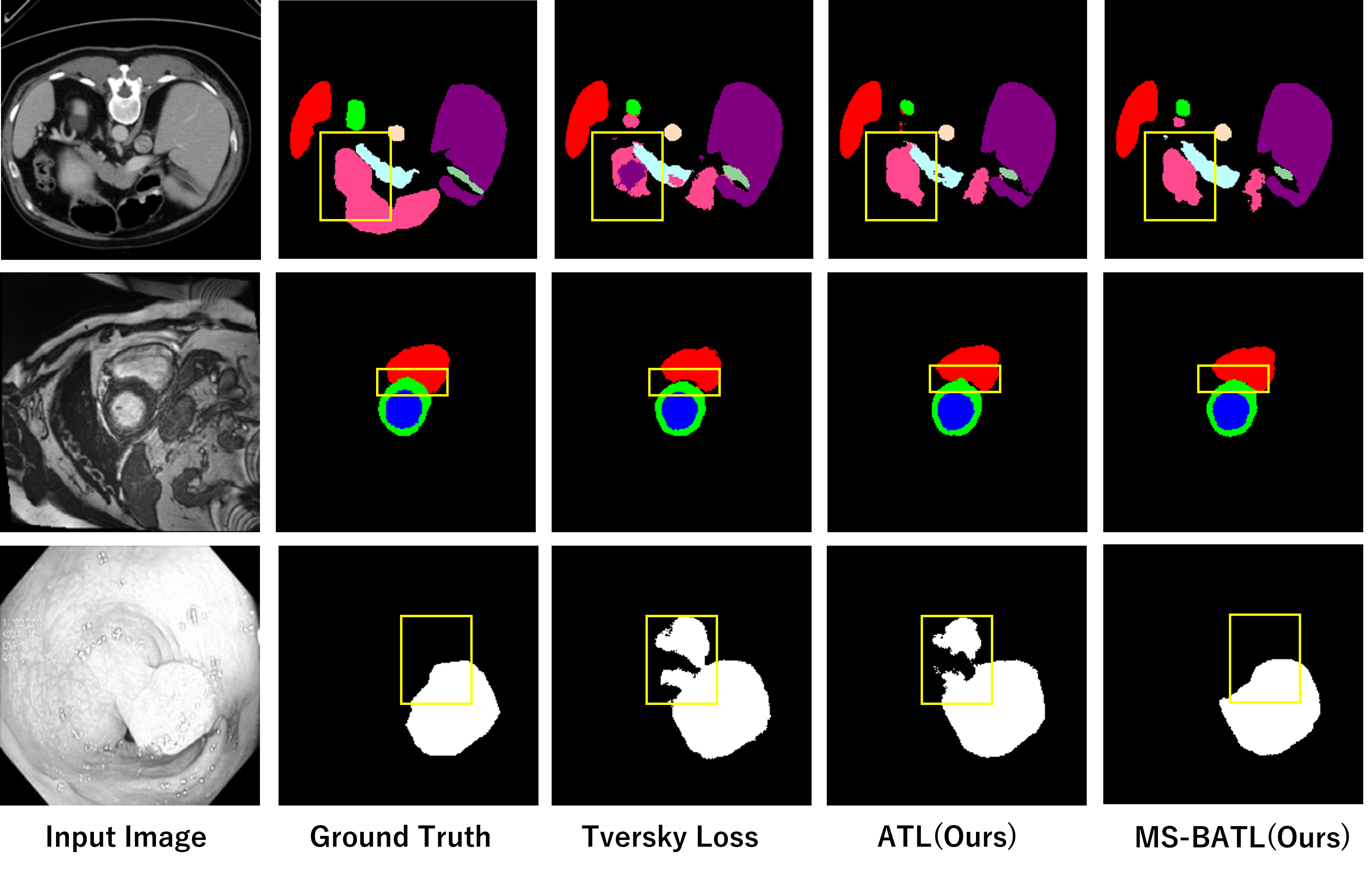}
        \centerline{(a) Qualitative segmentation results}
    \end{minipage}
    \hfill
    \begin{minipage}[t]{0.48\textwidth}
        \centering
        \includegraphics[width=\linewidth]{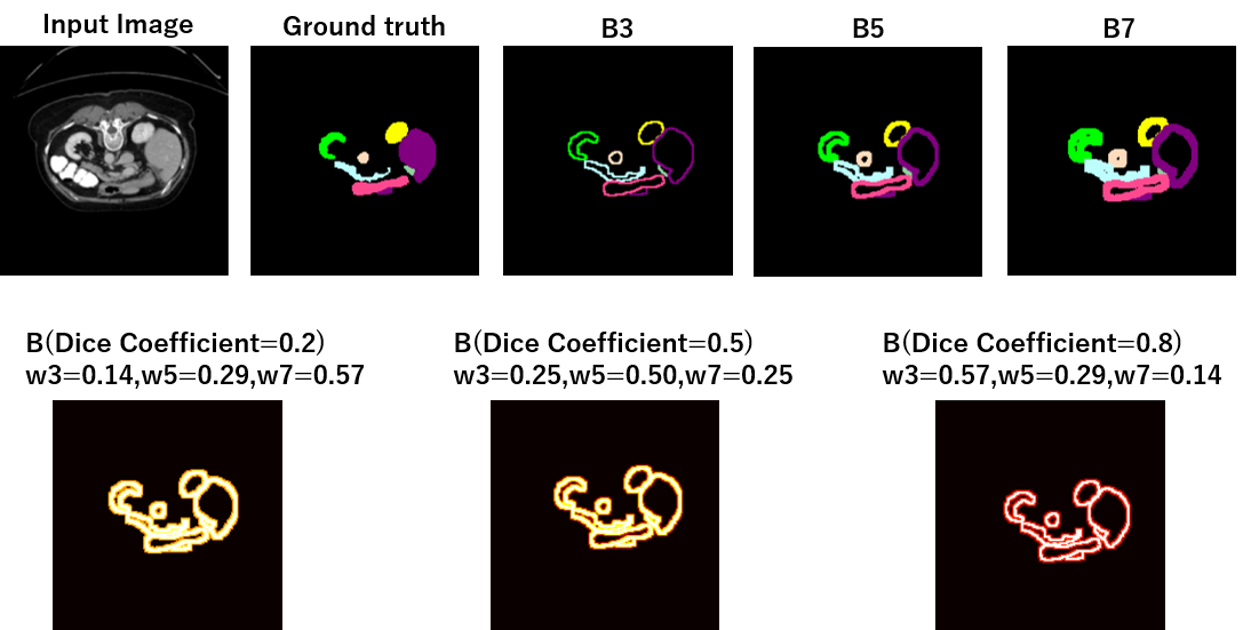}
        \centerline{(b) Boundary map visualization}
    \end{minipage}

    \caption{
    Qualitative segmentation results and boundary map visualization.
    (a) Qualitative results on Synapse multi-organ, ACDC, and Kvasir-SEG.
    From left to right, the columns show the input image, ground truth, Tversky Loss, ATL, and MS-BATL.
    Yellow boxes indicate regions with noticeable differences among methods.
    (b) Visualization of boundary maps in MS-BATL on the Synapse Multi-Organ dataset using U-Net.
    The top row shows the input image, ground-truth label, and boundary maps $B_3$, $B_5$, and $B_7$ generated using fixed $3 \times 3$, $5 \times 5$, and $7 \times 7$ kernels, respectively.
    The bottom row shows the integrated boundary maps for different Dice coefficients, where black indicates low boundary weights, red indicates moderate boundary weights, and yellow-white indicates high boundary weights.
    }
    \label{fig:vis_and_boundary}
\end{figure*}

\cref{fig:vis_and_boundary}(a) shows the qualitative evaluation results on the Synapse multi-organ, ACDC, and Kvasir-SEG datasets. Each row corresponds to Synapse, ACDC, and Kvasir-SEG from top to bottom, respectively. From left to right columns show the input image, ground truth, Tversky Loss with dataset-specific optimal weights, ATL, and MS-BATL. The yellow boxes indicate regions where noticeable differences among the methods can be observed.

For the Synapse multi-organ dataset, the background is shown in black, the aorta in light orange, the gallbladder in light green, the left kidney in green, the right kidney in yellow, the liver in purple, the pancreas in cyan, the spleen in red, and the stomach in pink. Within the yellow box, Tversky Loss shows missing regions and shape distortions in the stomach and pancreas regions. In contrast, ATL suppresses excessive fragmentation, while MS-BATL preserves the shapes around the stomach and pancreas more closely to the ground truth. This suggests that, in addition to the dynamic control of FP and FN by ATL, the multi-scale boundary information in MS-BATL may contribute to better boundary representation between organs.

To further illustrate how the boundary information is incorporated in MS-BATL, \cref{fig:vis_and_boundary} (b) visualizes the boundary maps used in MS-BATL on the Synapse Multi-Organ dataset using U-Net.
The top row shows the input image, ground truth, and boundary maps $B_3$, $B_5$, and $B_7$ generated with fixed $3 \times 3$, $5 \times 5$, and $7 \times 7$ kernels.
Larger kernels produce wider boundary bands, corresponding to coarser boundary information.
The bottom row shows integrated boundary maps for $D_{\mathrm{coefficient}}=0.2$, $0.5$, and $0.8$.
Brighter regions indicate larger boundary weights.
As the Dice coefficient increases, the emphasized regions shift from broad coarse boundaries to thinner fine boundaries, confirming the adaptive coarse-to-fine behavior of MS-BATL.
Returning to the qualitative results in \cref{fig:vis_and_boundary}(a), we then examine the ACDC and Kvasir-SEG examples.
For the ACDC dataset, the background is shown in black, the right ventricle in red, the myocardium in green, and the left ventricle in blue. Within the yellow box, Tversky Loss shows boundary misalignment between the right ventricle and myocardium, whereas ATL and MS-BATL produce spatial relationships that are closer to the ground truth. In particular, MS-BATL represents the contour of the myocardium more stably, suggesting that the incorporation of boundary information may help reduce errors around the boundary between the ventricle and myocardium.

For the Kvasir-SEG dataset, the background is shown in black and the polyp region in white. Within the yellow box, Tversky Loss and ATL show false-positive white regions that do not exist in the ground truth. In contrast, MS-BATL suppresses such false positives, and the predicted white region is concentrated closer to the ground-truth polyp region. This result suggests that MS-BATL may reduce false detections away from the target region by using boundary information as an auxiliary cue.

Overall, the qualitative results show that ATL tends to reduce over- and under-segmentation compared with Tversky Loss, while MS-BATL further suppresses boundary-related errors and false positives in some cases. However, since the effectiveness of MS-BATL depends on the dataset and target structure, it does not consistently improve performance under all conditions, as also observed in the quantitative evaluation.

\subsection{Ablation Study}

The ablation results are shown in \cref{tab:ablation}. For ATL, the proposed formulation in \cref{eq:atl_alpha_beta} achieved an mIoU of 59.94\%. In contrast, reversing $\alpha$ and $\beta$ substantially reduced the mIoU to 19.39\%. This indicates that emphasizing FN in the early stage of training is important for suppressing missed target regions.
We also evaluated a variant in which $\alpha$ was treated as a learnable parameter initialized to 0.0, while $\beta$ was computed as $\beta=1-\alpha$. This variant achieved an mIoU of only 0.40\%, which was 59.54\% lower than ATL. As discussed in \cref{sec:ATL}, directly optimizing the weighting coefficient can update it in a direction that is convenient for minimizing the training loss, rather than reflecting the current prediction performance. This result shows that simply making $\alpha$ learnable is insufficient, and that constraining the weights through the Dice coefficient is essential for stable weight control.
When $\alpha$ was directly set to the Dice coefficient, the mIoU was 57.04\%, which was 2.90\% lower than ATL. Direct Dice-based weighting can cause $\alpha$ to exceed $\beta$ once the Dice coefficient becomes larger than 0.5, leading to an early shift toward FP-oriented weighting. In contrast, ATL keeps $\alpha \leq \beta$ for $D_{\mathrm{coefficient}} \in [0,1]$ and gradually approaches $\alpha=\beta=0.5$, avoiding excessive FP-oriented weighting and promoting region-overlap-oriented optimization in the later stage.
For MS-BATL, the proposed method achieved an mIoU of 61.01\%, outperforming all ablation variants. When only a single boundary scale was used, the mIoU values for $3 \times 3$, $5 \times 5$, and $7 \times 7$ were 60.24\%, 59.04\%, and 60.53\%, respectively. These results indicate that a single boundary scale is insufficient and that combining multi-scale boundary information is effective. Although the $7 \times 7$ setting showed relatively high performance, it still underperformed MS-BATL, suggesting that integrating fine, intermediate, and coarse boundary information contributes to improve segmentation performance. Larger kernels such as $9 \times 9$ may produce excessively wide boundary bands, especially for small organs, and are left for future work.
Furthermore, removing the normalization in \cref{eq:multi_scale_boundary_map} decreased the mIoU to 59.89\%, which was 1.12\% lower than MS-BATL. This suggests that normalization is important for controlling the contribution of each scale and preventing excessive emphasis on boundary regions.
Overall, these ablation results demonstrate that the design direction of $\alpha$ and $\beta$ and the nonlinear Dice-coefficient-based weight update are important for ATL. They also confirm that normalized integration of multi-scale boundary information contributes to improved performance in MS-BATL.

\begin{table}[t]
\centering
\caption{
Ablation study of ATL and MS-BATL on the Synapse multi-organ dataset using U-Net.
$\Delta$ indicates the difference from the corresponding proposed method.
}
\label{tab:ablation}
\small
\begin{tabular}{llcc}
\toprule
Component & Variant & mIoU (\%) & $\Delta$ \\
\midrule

\multirow{4}{*}{ATL}
& ATL (Ours) & \textbf{59.94} & 0.00 \\
& Learnable $\alpha$ $(\alpha_{\mathrm{init}}=0.0,\ \beta=1-\alpha)$ & 0.40 & -59.54 \\
& Reversed weights $(\alpha \leftrightarrow \beta)$ & 19.39 & -40.55 \\
& Direct Dice-based weighting $(\alpha=D_{\mathrm{coefficient}})$ & 57.04 & -2.90 \\

\midrule

\multirow{5}{*}{MS-BATL}
& MS-BATL (Ours) & \textbf{61.01} & 0.00 \\
& Single-scale boundary $(3 \times 3)$ & 60.24 & -0.77 \\
& Single-scale boundary $(5 \times 5)$ & 59.04 & -1.97 \\
& Single-scale boundary $(7 \times 7)$ & 60.53 & -0.48 \\
& Without boundary-weight normalization & 59.89 & -1.12 \\

\bottomrule
\end{tabular}
\end{table}

\section{Conclusion}

In this paper, we proposed Adaptive Tversky Loss (ATL), which dynamically adjusts the weighting coefficients of Tversky Loss based on the Dice coefficient, and Multi-scale Boundary-Adaptive Tversky Loss (MS-BATL), which incorporates multi-scale boundary information into this adaptive framework. ATL reduces the need for manual tuning of FP and FN weights and enables prediction-performance-dependent control of the weighting balance.
Experimental evaluations across three medical image datasets showed that ATL achieves competitive performance without manually tuning the weighting coefficients. MS-BATL further improved performance in several multi-organ segmentation settings, suggesting that multi-scale boundary information can be beneficial when boundary cues are reliable.
The ablation study confirmed the importance of the nonlinear weighting formulation in ATL and the normalized multi-scale boundary integration in MS-BATL. Qualitative results also showed that the proposed methods can reduce over-segmentation and under-segmentation, and improve boundary delineation in challenging regions.
However, the effectiveness of the proposed methods depends on the model architecture and prediction stability, and consistent improvements were not observed in all cases. Future work will focus on more robust adaptive weighting strategies and more stable integration of boundary information.
Overall, the proposed adaptive Tversky-based framework serves as a practical and extensible approach for loss design in medical image segmentation, with the potential to improve generalizability and practical usability.

\bibliographystyle{splncs04}
\bibliography{main}
\end{document}